\documentclass[conference]{IEEEtran}
\ifCLASSINFOpdf
\else
\fi

\usepackage{graphicx}
\usepackage{listings}

\graphicspath{ {./Images/} }

\begin{document}
%
\title{ ASTM :\\Autonomous Smart Traffic Management System \\Using Artificial Intelligence CNN and LSTM}

\author{\IEEEauthorblockN{\textbf{Christofel Rio Goenawan}}
\IEEEauthorblockA{ Robotics Master Program\\
Korea Advanced Institute of Science and Technology\\
Daejeon, South Korea\\
Email: christofel.goenawan@kaist.ac.kr}
}
\maketitle

\begin{abstract}

In the modern world, the development of Artificial Intelligence (AI) has contributed to improvements in various areas, including automation, computer vision, fraud detection, and more. AI can be leveraged to enhance the efficiency of Autonomous Smart Traffic Management (ASTM) systems and reduce traffic congestion rates. This paper presents an Autonomous Smart Traffic Management (STM) system that uses AI to improve traffic flow rates. The system employs the YOLO V5 Convolutional Neural Network to detect vehicles in traffic management images. Additionally, it predicts the number of vehicles for the next 12 hours using a Recurrent Neural Network with Long Short-Term Memory (RNN-LSTM). The Smart Traffic Management Cycle Length Analysis manages the traffic cycle length based on these vehicle predictions, aided by AI.
From the results of the RNN-LSTM model for predicting vehicle numbers over the next 12 hours, we observe that the model predicts traffic with a Mean Squared Error (MSE) of 4.521 vehicles and a Root Mean Squared Error (RMSE) of 2.232 vehicles. After simulating the STM system in the CARLA simulation environment, we found that the Traffic Management Congestion Flow Rate with ASTM (21 vehicles per minute) is 50\% higher than the rate without STM (around 15 vehicles per minute). Additionally, the Traffic Management Vehicle Pass Delay with STM (5 seconds per vehicle) is 70\% lower than without ASTM (around 12 seconds per vehicle). These results demonstrate that the ASTM system using AI can increase traffic flow by 50\% and reduce vehicle pass delays by 70\%.

\textit{Keywords : Smart Traffic Management System , Automation, Artificial Intelligence, Computer Vision, Recurrent Neural Network, Traffic Simulation}
\end{abstract}


%
\IEEEpeerreviewmaketitle

\section{Introduction}

\subsection{ Artificial Intelligence}
The development of Artificial Intelligence (AI) began in 1943 when neurophysiologist Warren McCulloch and mathematician Walter Pitts published a paper introducing Artificial Neural Networks (ANN) to the world \cite{anyoha2017history}. The development of AI started to gain attention when British polymath Alan Turing published his paper "Can Machines Think?" in 1950, where Turing suggested that machines can do the same things as humans, using available information and reasoning to solve problems and make decisions \cite{anyoha2017history}. The development of AI officially started after Allen Newell, Cliff Shaw, and Herbert Simon published the \textit{proof of concept} in the first AI program named \textit{Logic Theorist} in 1955 \cite{anyoha2017history}.

After that, until 1974, AI grew rapidly because a lot of investment was put into the field, and computers could store more information and became faster, cheaper, and more accessible. However, at the start of 1980, AI entered the "dark era" due to many obstacles, such as a lack of computational power to do anything substantial; computers simply couldn’t store enough information or process it fast enough, which discouraged many investors and researchers from delving deeper into this field. Until the end of the 20th century, AI development went through a roller coaster of success and setbacks, including famous “deep learning” techniques that allowed computers to learn using experience, popularized by John Hopfield and David Rumelhart in the 1980s \cite{schmidhuber2015deep}.

In 1997, AI regained hype after IBM’s Deep Blue, a chess-playing computer program, defeated reigning world chess champion and grandmaster Garry Kasparov \cite{anyoha2017history}. However, it was not until 2012 that AlexNet, a Convolutional Neural Network (CNN) architecture by Alex Krizhevsky and his team, won the 2012 ImageNet annual image recognition challenge by a huge margin, reigniting development in the AI field \cite{geron2017hands}. To this day, numerous developments have emerged in AI, attributed to the vast amount of data available to train models, tremendous increases in computing power since the 1990s, and the introduction of revolutionary AI architectures with significantly higher performance and usability.

Nowadays, Artificial Intelligence (AI) development has contributed to improvements in various areas, including scientific research, industry, environmental sectors, and governmental and social issues. For example, AI has proven effective in solving a variety of practical problems such as disease detection \cite{esteva2017dermatologist}, language translation \cite{wu2016google}, autonomous self-driving cars \cite{goenawan2024unseen}, and customer behavior prediction \cite{wang2018webpage}.

However, AI development has been hampered by difficulties in sharing ML models and differences in dependencies and machine environments, making it challenging to deploy a model across different machines \cite{linux2020acumos}. Usually, ML models contain multi-stage, complex pipelines with procedures that are sequentially entangled and mixed together, such as preprocessing, feature extraction, data transformation, training, and validation \cite{geron2017hands}. Hence, improving an individual component may, in fact, worsen overall performance due to the strong correlations between components. Therefore, building models becomes a trial-and-error-based iterative process that demands expert-level knowledge in ML concepts to create and tune ML models manually \cite{sculley2015technical}. Moreover, because of the dependencies for different AI tools like TensorFlow, Keras, and PyTorch, machine dependencies must be installed manually, which is often a time-wasting process and not always straightforward \cite{ribeiro2015mlaas}.

\subsection{ Object Detection using Computer Vision }
Object Detection using Computer Vision involves detecting objects around Artificial Intelligence Sensor Image Cameras. In computer vision, convolutional neural networks (CNNs) are very popular for tasks like image classification, object detection, image segmentation, drivable area detection \cite{goenawan2024unseen}, 3D point cloud object completion \cite{goenawan2024_enhancing_point_completion_network} and more. Image classification is one of the most needed techniques in today’s era; it is used across various domains like healthcare and business. Thus, knowing and creating your own state-of-the-art computer vision model is a must if you’re in the AI domain. Most computer vision algorithms utilize something called a convolutional neural network (CNN). A CNN is a model used in machine learning to extract features, like texture and edges, from spatial data. Like basic feedforward neural networks, CNNs learn from inputs, adjusting their parameters (weights and biases) to make accurate predictions. However, what makes CNNs special is their ability to extract features from images. Take an image of a car, for example. In a normal feedforward neural network, the image would be flattened into a feature vector. However, CNNs can treat images like matrices and extract spatial features, such as texture, edges, and depth. They accomplish this through convolutional layers and pooling. The architecture of the Artificial Intelligence Convolutional Neural Network can be seen below.

\begin{figure}[h]
    \centering
    \includegraphics[width=8cm,scale=1]{ 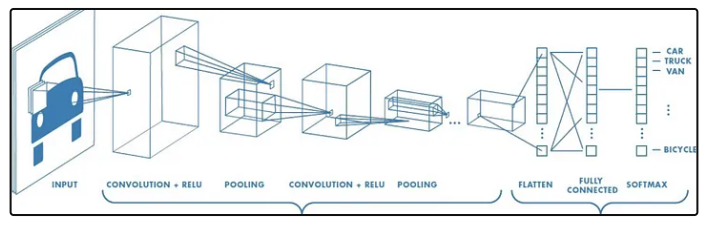}
    \caption{Convolutional Neural Network for Image Detection in Artificial Intelligence}
\end{figure}

The CNN processes an image by applying a series of filters, resulting in feature maps that highlight different aspects of the input image. The convolution kernel, which is a matrix of weights, slides over the input image matrix, performing element-wise multiplications and summing the results to produce a feature map. These feature maps allow CNNs to understand and categorize images based on spatial features. 

\subsection{ Traffic Congestion Prediction using Artificial Intelligence }
One application of Artificial Intelligence is predicting traffic congestion. By analyzing traffic time, traffic jam conditions, and weather conditions, AI can predict traffic congestion effectively. Recurrent neural networks (RNNs) are deep learning models typically used to solve problems with sequential input data, such as time series. RNNs retain a memory of previously processed inputs and learn from these iterations during training \cite{linux2020acumos}. 

To understand RNNs, consider that they are a class of artificial neural networks where connections between nodes form a directed graph along a temporal sequence. This allows them to exhibit temporal dynamic behavior. Unlike feedforward neural networks, which do not retain memory, RNNs can process variable-length sequences of inputs \cite{wang2018webpage}. RNNs share parameters across each layer of the network, and while they adjust weights during training, they often face challenges such as exploding and vanishing gradients \cite{vanrijn2013openml}. The architecture of the Recurrent Neural Network for predicting traffic congestion can be seen below.

\begin{figure}[h]
    \centering
    \includegraphics[width=8cm,scale=1]{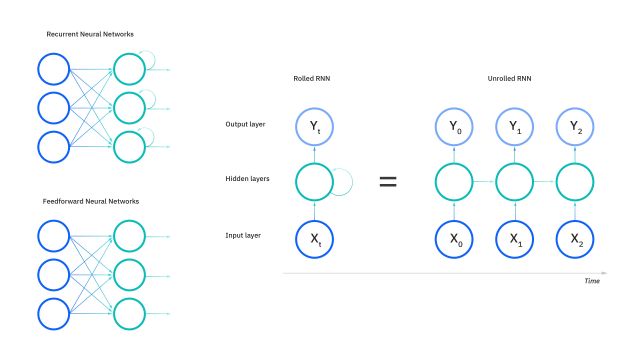}
    \caption{Recurrent Neural Network Architecture for Traffic Congestion System Prediction using Artificial Intelligence}
\end{figure}

\subsection{ Smart Traffic Management Systems }
One cornerstone of smart city design is having an integrated smart transportation solution. It can be argued that a city is not completely intelligent without a smart traffic management system. Intelligent transportation systems (ITS) or smart traffic management systems (Figure 1) provide an organized, integrated approach to minimizing congestion and improving safety on city streets through connected technology. The intelligent traffic management system market is expected to grow to \$19.91 billion by 2028 at a 10.1\% CAGR, according to PR Newswire. The demand and increased adoption rate of smart traffic management solutions can be attributed to the rise of smart city technology. Guidehouse Insights reports that there are more than 250 smart city projects globally. 

Symmetry Electronics supplier, Digi International, defines smart traffic management systems as technology solutions that municipalities can integrate into their traffic cabinets and intersections today for fast, cost-effective improvements in safety and traffic flow on their city streets. Efficient and successful smart traffic management systems utilize next-generation hardware and software to optimize traffic infrastructure (Figure 2). The architecture of the Smart Traffic Management System using Artificial Intelligence can be seen below.

\begin{figure}[h]
    \centering
    \includegraphics[width=8cm,scale=1]{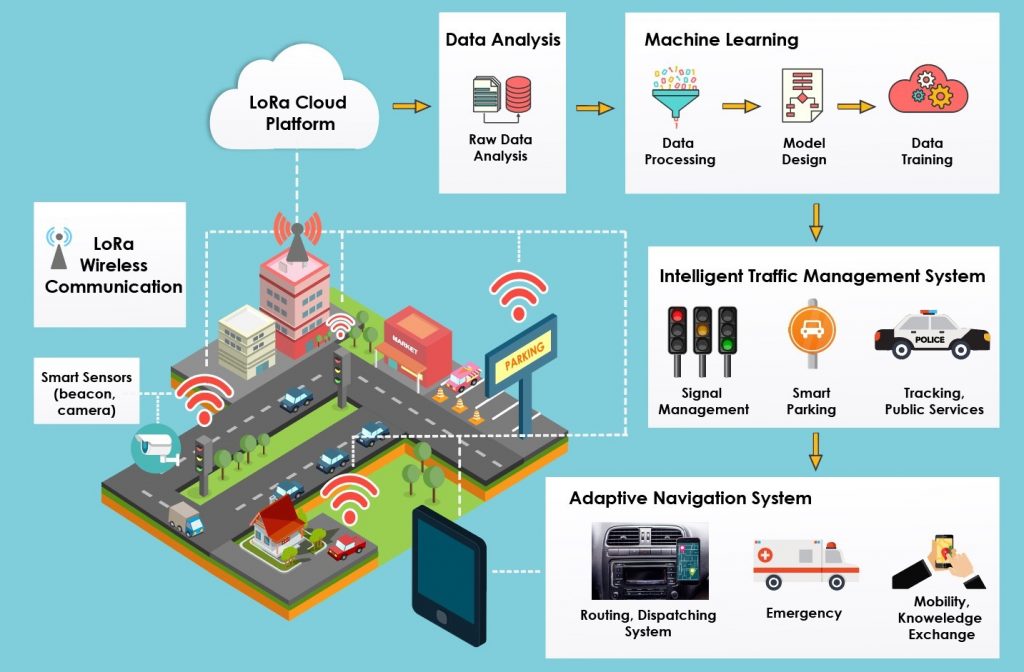}
    \caption{Architecture of Smart Traffic Management using Artificial Intelligence}
\end{figure}

Transportation plays a crucial role in any community, connecting people to jobs, services, and opportunities. Monitoring key metrics can provide valuable insight into how a transportation system is performing and whether it meets the needs of those who rely on it. This section explores some key transportation metrics to analyze in any community, including measures of mobility, road safety, and accessibility. By understanding these metrics, community leaders and transportation professionals can make informed decisions to improve transportation systems and enhance the overall quality of life for residents. Metrics to Measure Traffic Congestion in Smart Traffic Management can be seen below.

\begin{enumerate}
    \item Average Daily Traffic (ADT) and Annual Average Daily Traffic (AADT): ADT and AADT quantify how busy a stretch of road or highway is, reporting the number of vehicles passing through over a day or year, respectively. They have many applications within traffic engineering, such as signal timing and determining where infrastructure investments should go. 

    \item Corridor Travel Times: Corridor Travel Times help transportation agencies understand how long it takes to travel between two points and allow them to identify bottlenecks and improve planning and programming. 

    \item Speed: A common measure of traffic congestion is the speed at which vehicles travel on a roadway. It can be averaged over specific time intervals or collected for an entire day, providing insight into how traffic conditions change throughout the day. 

    \item Travel Time Index (TTI): TTI is a ratio of the travel time during peak hours to the travel time during free-flow conditions. TTI identifies congestion and can help agencies determine the effectiveness of congestion management strategies.

    \item Delay: Delay is the difference between the time a vehicle would take to travel in free-flow conditions and the time taken to travel under congested conditions. It helps identify congestion levels and improve roadway operations.

    \item Level of Service (LOS): LOS is a grading system from A to F used to evaluate traffic flow, where A indicates free-flow conditions and F indicates heavy congestion. 

    \item Peak Period: The peak period is the time when congestion is at its highest level, usually occurring during morning and evening rush hours. Identifying peak periods can help agencies optimize traffic management strategies.
\end{enumerate}

It is essential to analyze metrics to identify existing issues and prioritize solutions. Advanced traffic management systems can continuously collect data from the street network and dynamically respond to changing traffic conditions. Traffic signal timing can be adapted based on current congestion levels, and information can be provided to drivers in real time through digital signs or mobile applications. As technology continues to advance, the capabilities of smart traffic management systems will continue to expand, ultimately leading to safer, more efficient roadways.

\section{Methodology}

In this project, the author proposes a novel architecture for a Smart Traffic Management System by using Artificial Intelligence to intelligently manage traffic. First, we use Artificial Intelligence to detect vehicles in traffic management system images. After detecting vehicles, we predict the number of vehicles using Artificial Intelligence to estimate the traffic density. Then, the system manages the traffic cycle length based on fuzzy decision-making for traffic flow. The architecture of the autonomous Smart Traffic Management System using Artificial Intelligence is illustrated in Figure 4.

\begin{figure*}[h]
    \centering
  \includegraphics[width=\textwidth,height=8cm]{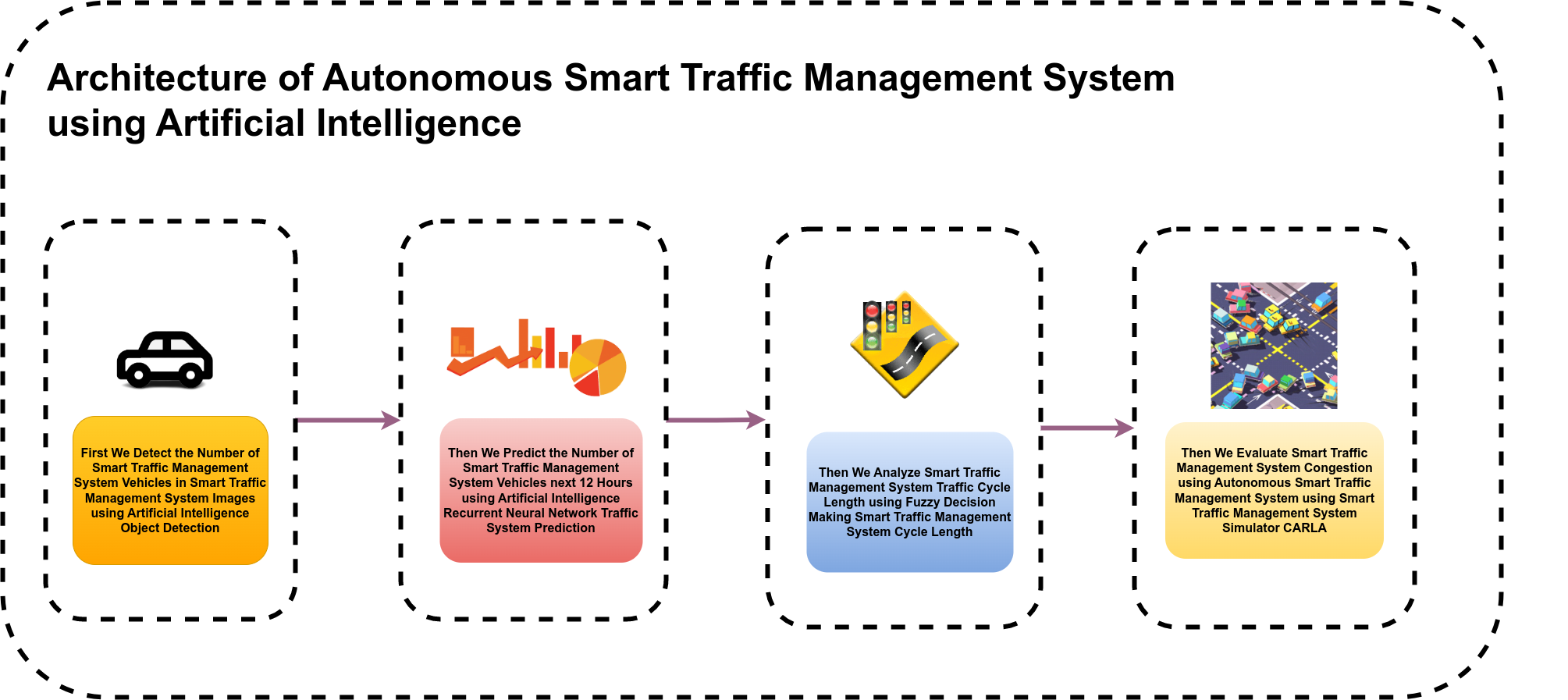}
  \caption{Architecture of Autonomous Smart Traffic Management System using Artificial Intelligence}
\end{figure*}

\subsection{Smart Traffic Management System Vehicle Detection using Convolutional Neural Network}

First, we detect vehicles in traffic management system images using a Convolutional Neural Network (CNN). The CNN detects vehicles by analyzing the pixels and colors in the images. For vehicle detection, the author uses the \textit{Road Vehicle Image Dataset of Bangladesh Traffic}, created by Ashfak Yeafi in 2017. This dataset contains annotated images of Bangladesh road vehicles, with separate folders for training and validation images. The dataset includes over a million images, captured from January to December 2017.

For the vehicle detection task, the author uses the YOLOv5 CNN model. YOLOv5 is widely used for object detection tasks and comes in four versions: small (s), medium (m), large (l), and extra-large (x), each offering progressively higher accuracy. YOLOv5 works by extracting features from images and predicting bounding boxes and class labels for detected objects. It connects the process of predicting bounding boxes and class labels in an end-to-end differentiable network.

The YOLO network consists of three main components:

\begin{enumerate}
\item \textbf{Backbone}: A convolutional neural network that aggregates and forms image features at different granularities.
\item \textbf{Neck}: A series of layers to mix and combine image features before passing them forward to prediction.
\item \textbf{Head}: Consumes features from the neck and performs the final box and class prediction.
\end{enumerate}

The architecture of YOLOv5 for Smart Traffic Management System vehicle detection using CNN is shown below.

\begin{figure}[h]
    \centering
    \includegraphics[width=8cm,scale=1]{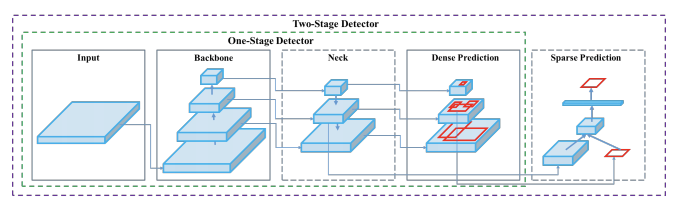}
    \caption{Architecture of YOLO V5 Smart Traffic Management System Vehicles Detection using Convolutional Neural Network}
\end{figure}

During the vehicle detection process, the author tunes five key hyperparameters of the YOLOv5 model. These hyperparameters are:

\begin{enumerate}
    \item \textbf{Learning-rate start (lr0)}: Determines the step size at each iteration. For example, a learning rate of 0.1 means the training progress increases by 0.1 at each iteration.
    \item \textbf{Momentum}: A parameter for the gradient descent algorithm that replaces the gradient with an aggregate of gradients.
    \item \textbf{Mosaic}: Increases model accuracy by creating a new image from multiple combined images for data augmentation.
    \item \textbf{Degree}: Improves model accuracy by randomly rotating images during training, up to 360 degrees.
    \item \textbf{Scaling}: Resizes images to either match grid size or optimize results.
    \item \textbf{Weight Decay}: A regularization technique that penalizes large weights to prevent overfitting.
\end{enumerate}

The six most important hyperparameters for YOLOv5 tuning in vehicle detection can be seen in the following figure.

\begin{figure}[h]
    \centering
    \includegraphics[width=8cm,scale=1]{ 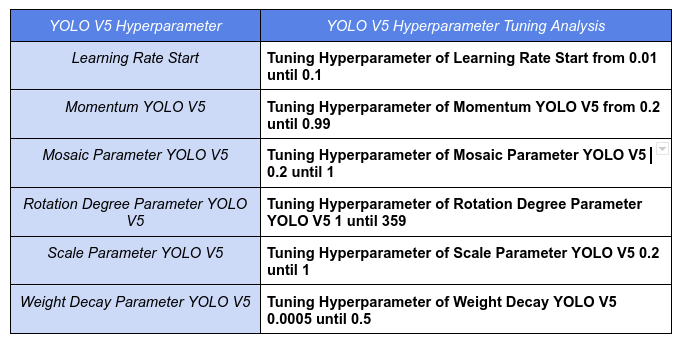}
\end{figure}

\subsection{Traffic Vehicle Prediction using Recurrent Neural Network}

After determining the number of vehicles, the author predicts the next 12 hours of traffic using a Recurrent Neural Network (RNN). Specifically, a Long Short-Term Memory (LSTM) model is used to predict future traffic flow. LSTMs are designed to handle long-term dependencies in sequential data, making them suitable for tasks like time series forecasting.

An LSTM has a memory cell controlled by three gates: the input gate, the forget gate, and the output gate. These gates manage the flow of information, allowing the LSTM to selectively retain important data over time. This makes LSTMs ideal for traffic prediction tasks.

The architecture of the RNN with LSTM for predicting traffic in a Smart Traffic Management System is shown in Figure 7.

\begin{figure}[h]
    \centering
    \includegraphics[width=8cm,scale=1]{ 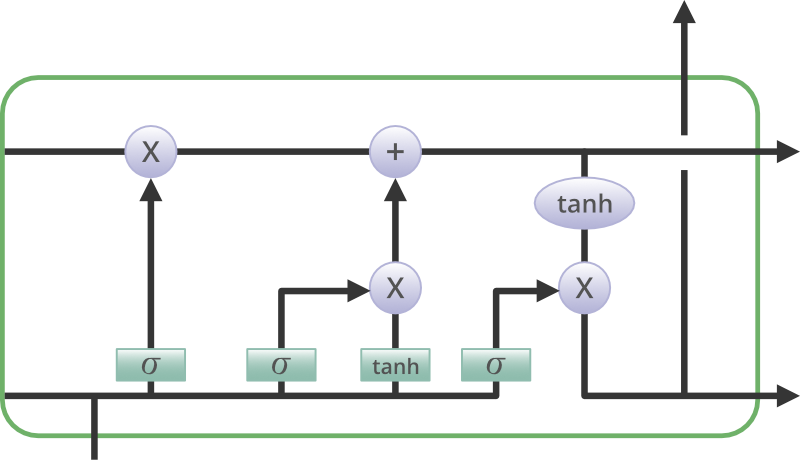}
    \caption{Architecture of Recurrent Neural Network Long Short Term Memory for Predicting Traffic Vehicles in Smart Traffic Management System}
\end{figure}

The author uses several features for predicting the number of vehicles over the next 12 hours. These features are illustrated in Figure 8.

\begin{figure}[h]
    \centering
    \includegraphics[width=8cm,scale=1]{ 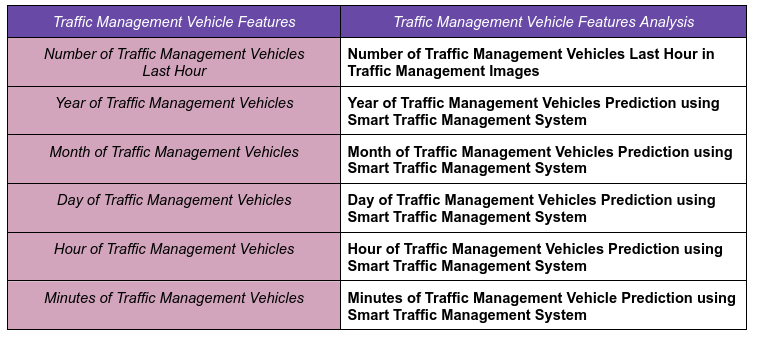}
\end{figure}

\subsection{Smart Traffic Management Traffic Cycle Length Analysis using Fuzzy Decision Making}

The Smart Traffic Management Traffic Cycle Length is managed based on vehicle prediction using Artificial Intelligence. The traffic cycle length is the total signal time required to serve all signal phases, including green time and any change intervals. Longer cycles accommodate more vehicles per hour but also produce higher average delays.

The optimal cycle length can be determined using Webster's formula, which minimizes intersection delay:

$$ C = \frac{(1.5 \times L + 5)}{(1.0 - SY_i)} $$

Where:
$C$ = optimal cycle length in seconds (rounded to the nearest 5 seconds),
$L$ = unusable time per cycle (e.g., signal change intervals),
$SY_i$ = critical lane volume for each phase/saturation flow.

The saturation flow is typically between 1500 and 1800 vehicles per hour. This formula can be used to calculate the signal timing for planning purposes. After determining the cycle length, the green time can be proportioned for each phase based on critical lane volumes.

\subsection{Evaluating the Smart Traffic Management System using Artificial Intelligence with the Traffic Management Simulator CARLA}

After determining the Smart Traffic Management System Traffic Cycle Length using Smart Traffic Cycle Length Fuzzy Decision Making, the user will simulate the Smart Traffic Management System using Artificial Intelligence in the Smart Traffic Management Simulator CARLA. CARLA is an open-source autonomous driving simulator. It was built from scratch to serve as a modular and flexible API to address a range of tasks involved in the problem of autonomous driving. One of the main goals of CARLA is to help democratize autonomous driving R\&D, serving as a tool that can be easily accessed and customized by users. To do so, the simulator must meet the requirements of different use cases within the general problem of driving (e.g., learning driving policies, training perception algorithms, etc.). CARLA is based on Unreal Engine to run the simulation and uses the OpenDRIVE standard (1.4 as of today) to define roads and urban settings. Control over the simulation is granted through an API handled in Python and C++ that is constantly being improved as the project evolves. In order to smooth the process of developing, training, and validating driving systems, CARLA has evolved into an ecosystem of projects, built around the main platform by the community. In this context, it is important to understand how CARLA works to fully comprehend its capabilities.


First, the author tunes the hyperparameters of the YOLO V5 Convolutional Neural Network to detect Smart Traffic Management System vehicles. The tuning process for the YOLO V5 Convolutional Neural Network hyperparameters for detecting vehicles in the Smart Traffic Management System is shown in Figure 8.

\begin{figure}[h] \centering \includegraphics[width=8cm,scale=1]{ 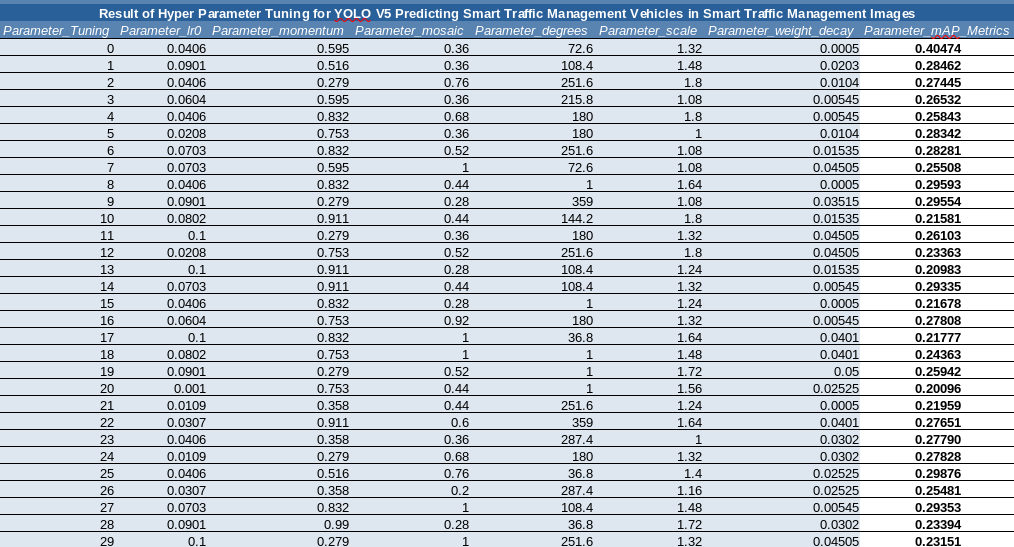} \caption{ Result Table of Tuning Hyperparameter YOLO V5 to Predict Smart Traffic Management System Vehicles in Smart Traffic Management Images } \end{figure}

From the results of tuning the hyperparameters of the YOLO V5 Convolutional Neural Network to detect vehicles in the Smart Traffic Management System, we can see that the best hyperparameter settings are from Search 1 with a learning rate of 0.0052, momentum of 0.595, mosaic parameter of 0.36, rotation degree of 72, scale of 1.32, and weight decay of 0.0005, achieving a mean average precision (MAP) of 0.4034, which is almost 50

\begin{figure}[h] \centering \includegraphics[width=8cm,scale=1]{ 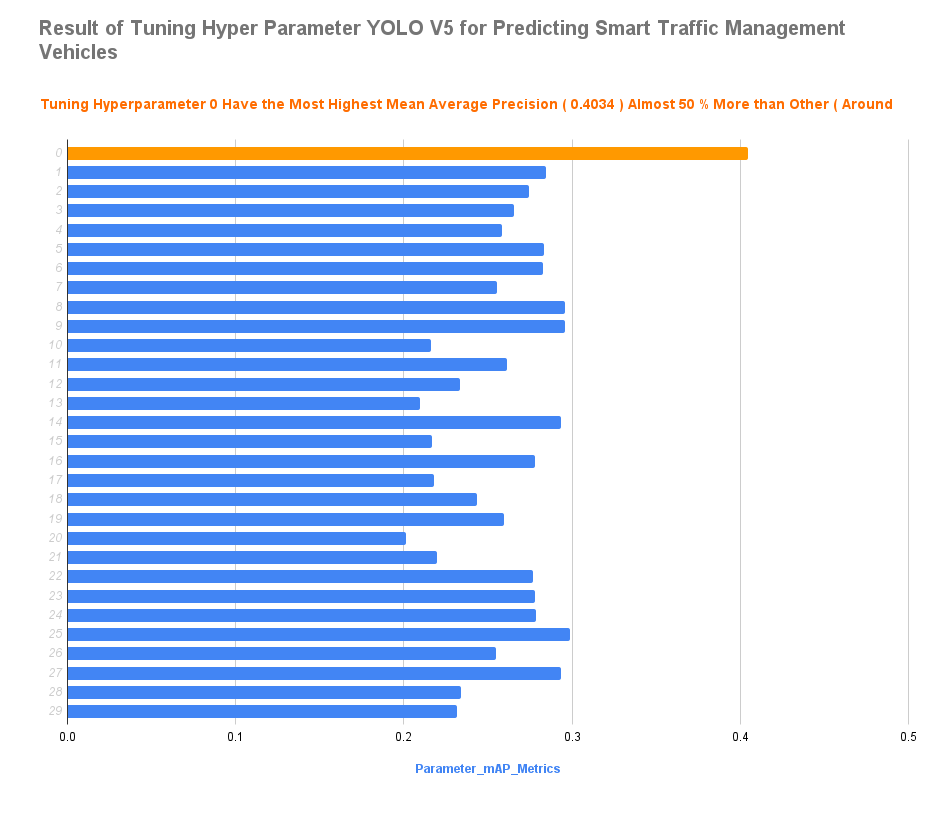} \caption{ Results of Best Tuning Hyperparameters for YOLO V5 to Predict Smart Traffic Management System Vehicles in Smart Traffic Management Images} \end{figure}

\begin{figure*}[h] \centering \includegraphics[width=\textwidth,height=8cm]{ 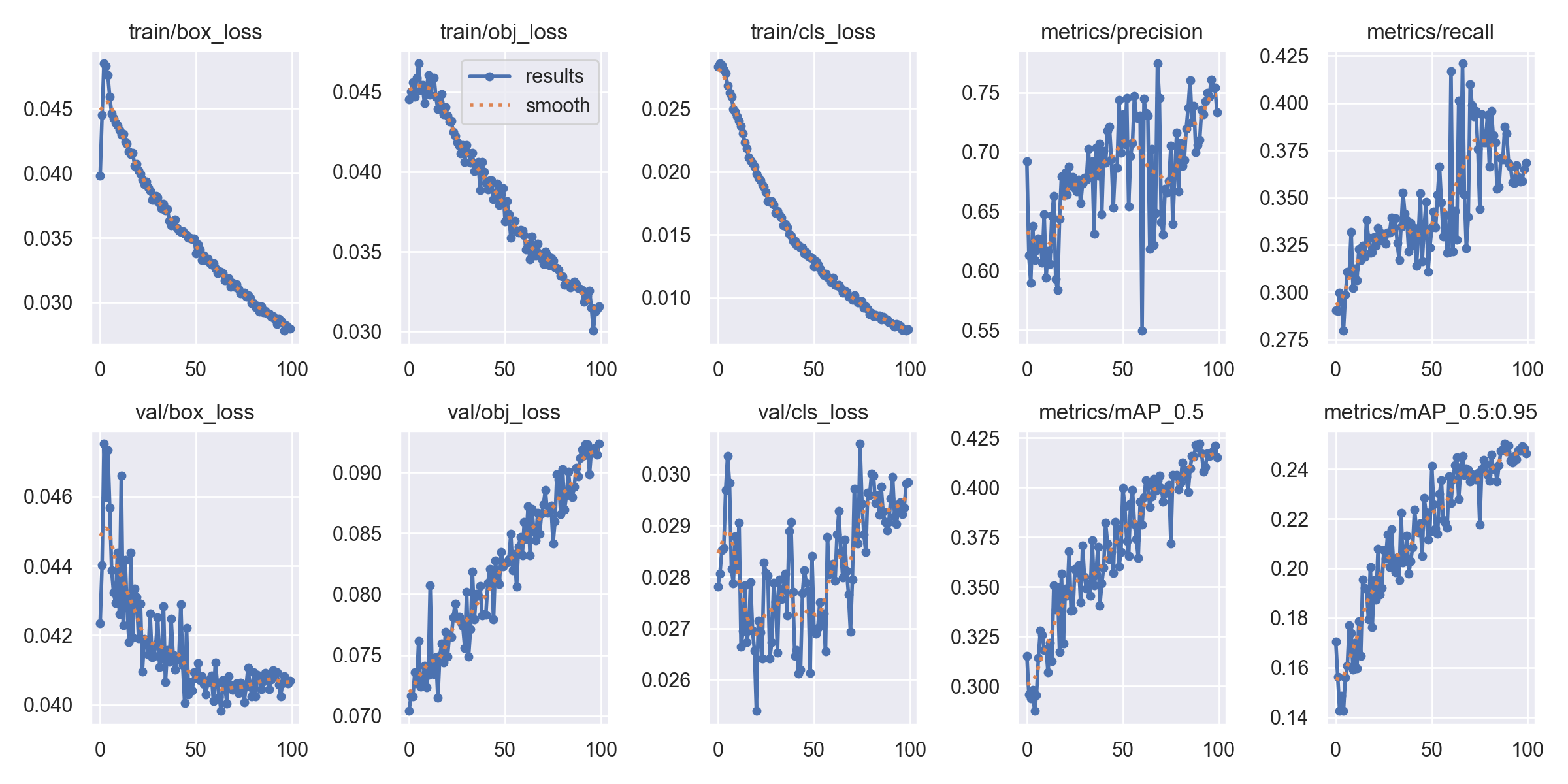} \caption{ Results of Training the Best Tuning Hyperparameter for the YOLO V5 Convolutional Neural Network to Detect Smart Traffic Management Vehicles in Smart Traffic Management Images} \end{figure*}

\begin{figure}[h] \centering \includegraphics[width=8cm,scale=1]{ 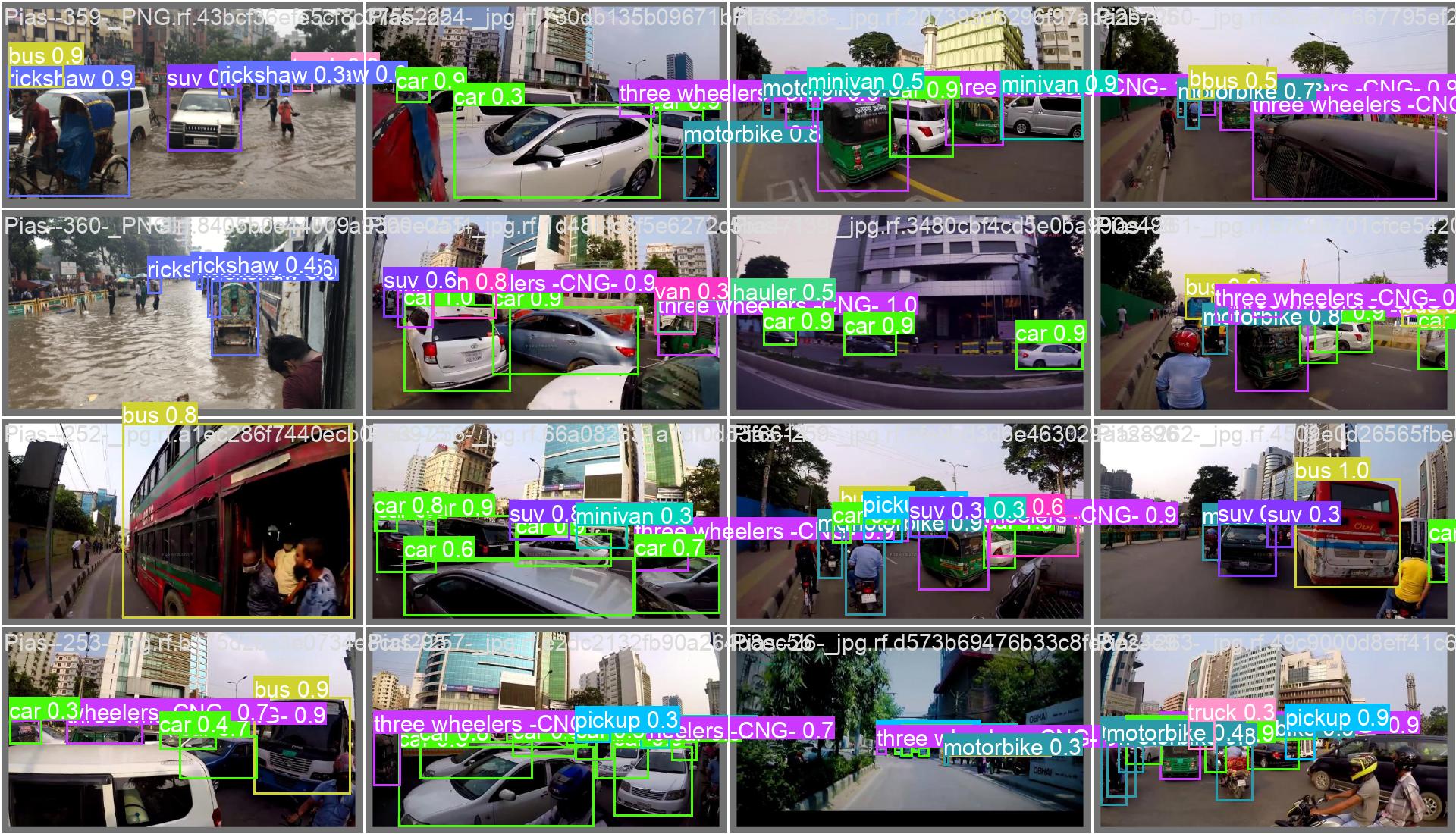 } \caption{ YOLO V5 Convolutional Neural Network Predicts Smart Traffic Management Vehicles in Smart Traffic Management Images} \end{figure}

Next, the author trains the YOLO V5 Convolutional Neural Network with the best-tuned hyperparameters to detect Smart Traffic Management Vehicles in Smart Traffic Management Images. The best hyperparameter settings used for training include a learning rate of 0.00406, momentum of 0.595, mosaic parameter of 0.36, rotation degree of 72.6, scale of 1.32, and weight decay of 0.0005. The results of training the YOLO V5 Convolutional Neural Network with the best-tuned hyperparameters for detecting vehicles are shown in Figure 9.

From the training process, we see that the YOLO V5 Convolutional Neural Network with the best-tuned hyperparameters can detect Smart Traffic Management Vehicles in Smart Traffic Management Images with a mean average precision (MAP) of 0.88561. For detecting cars specifically, the MAP is 0.95251. Additionally, the training loss decreased to 0.0010, which is 50

Next, the author trains a Recurrent Neural Network Long Short-Term Memory (RNN-LSTM) to predict Smart Traffic Management Vehicles. The RNN-LSTM is trained to predict the number of vehicles over the next 12 hours using features such as the number of vehicles in the last hour, year, month, day, hour, and minute. The results of the RNN-LSTM for predicting vehicles over the next 12 hours are shown below.

\begin{figure}[h] \centering \includegraphics[width=8cm,scale=1]{ 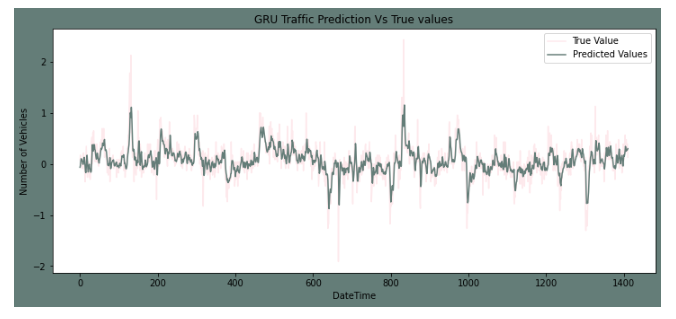 } \caption{ Results of Recurrent Neural Network Long Short-Term Memory Predictions for Smart Traffic Management Vehicles in the Next 12 Hours} \end{figure}

From the results of the RNN-LSTM model, we can see that it predicts the number of vehicles in the next 12 hours with a mean squared error (MSE) of 4.521 vehicles and a root mean squared error (RMSE) of 2.232 vehicles. The model can accurately predict both increases and decreases in the number of vehicles over the next 12 hours.

The author then evaluates the Smart Traffic Management System using Artificial Intelligence in the CARLA Traffic Management Simulator. A total of 100 scenarios are simulated each day, comparing the performance of the Smart Traffic Management System using AI and a conventional Traffic Management System. The evaluation results are shown below.

\begin{figure}[h] \centering \includegraphics[width=8cm,scale=1]{ 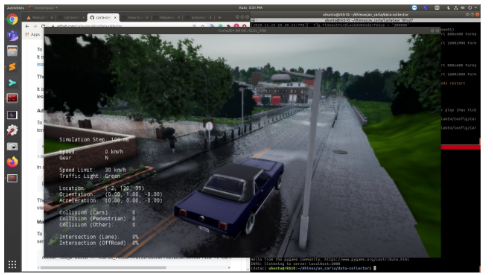} \caption{ Evaluation Results of the Smart Traffic Management System using Artificial Intelligence in the CARLA Traffic Management Simulator} \end{figure}

After simulating the Smart Traffic Management System in CARLA, we observe that the Traffic Management Congestion Flow Rate with the Smart Traffic Management System (21 vehicles per minute) is 50

\section{Conclusion}

This paper presents an Autonomous Smart Traffic Management System using Artificial Intelligence to improve traffic congestion flow rates. The system utilizes the YOLO V5 Convolutional Neural Network to detect vehicles in traffic management images. The hyperparameters of the YOLO V5 model, including the learning rate, momentum, mosaic parameter, rotation degree, scale, and weight decay, were optimized. The best hyperparameter settings resulted in a mean average precision (MAP) of 0.4034, which is almost 50\% higher than other configurations (approximately 0.2852). After training the model, the YOLO V5 Convolutional Neural Network achieved an MAP of 0.88561, and for cars specifically, the MAP was 0.95251.

The Smart Traffic Management System also predicts the number of vehicles over the next 12 hours using a Recurrent Neural Network Long Short-Term Memory (RNN-LSTM). The model predicts vehicle numbers with an MSE of 4.521 and an RMSE of 2.232, accurately predicting traffic trends for the next 12 hours.

Finally, the Smart Traffic Management System is evaluated using the CARLA simulator. The results show that the Traffic Management Congestion Flow Rate with the Smart Traffic Management System (21 vehicles per minute) is 50\% higher than without it (around 15 vehicles per minute), and the Traffic Management Vehicle Pass Delay is 70\% less with the system (5 seconds per vehicle compared to around 12 seconds). These findings highlight the effectiveness of the Smart Traffic Management System using AI in reducing traffic congestion and vehicle delay.

\section*{List of Abbreviation}
\begin{itemize}

\item ASTM \quad: Autonomous Smart Traffic Management
\item AI \quad: Artificial Intelligence
\item ANN \quad: Artificial Neural Network
\item CNN \quad: Convolutional Neural Network
\item RNN \quad: Recurrent Neural Network
\item LSTM \quad: Long Short Term Memory
\item RMSE \quad: Root Mean Squared Error
\end{itemize}



\section*{Declaration of competing interest}

The authors declare that they have no known competing financial interests or personal relationships that could have appeared to influence the work reported in this paper.

\section*{Data availability}
Data and experiment code and result will be made available on request.

\pagebreak



%

\end{document}